\documentclass{bmvc2k}

\usepackage{graphicx}
\usepackage{booktabs}
\usepackage{amsmath}
\usepackage{amssymb}
\usepackage{diagbox}
\usepackage{array}

\title{Addressing Data Bias Problems for Chest X-ray Image Report Generation}

\addauthor{Philipp Harzig}{philipp.harzig@informatik.uni-augsburg.de}{1}
\addauthor{Yan-Ying Chen}{yanying@fxpal.com}{2}
\addauthor{Francine Chen}{chen@fxpal.com}{2}
\addauthor{Rainer Lienhart}{rainer.lienhart@informatik.uni-augsburg.de}{1}

\addinstitution{
 Multimedia Computing and Computer Vision Lab\\
 University of Augsburg\\
 Augsburg, Germany
}
\addinstitution{
 FX Palo Alto Laboratory\\
 3174 Porter Drive,\\
 Palo Alto, CA, USA
}

\runninghead{Harzig et al.}{Data Bias Problems for CXR Medical Report Generation}

\newcolumntype{C}[1]{>{\centering\let\newline\\\arraybackslash\hspace{0pt}}m{#1}}
\newcolumntype{L}[1]{>{\raggedright\let\newline\\\arraybackslash\hspace{0pt}}m{#1}}

\begin{document}

\maketitle

\begin{abstract}
Automatic medical report generation from chest X-ray images is one possibility for assisting doctors to reduce their workload. 
However, the different patterns and data distribution of normal and abnormal cases can bias machine learning models.
Previous attempts did not focus on isolating the generation of the abnormal and normal sentences in order to increase the variability of generated paragraphs.
To address this, we propose to separate abnormal and normal sentence generation by using a dual word LSTM in a hierarchical LSTM model.
In addition, we conduct an analysis on the distinctiveness of generated sentences compared to the BLEU score, which increases when less distinct reports are generated. Together with this analysis, we propose a way of selecting a model that generates more distinctive sentences. We hope our findings will help to encourage the development of new metrics to better verify methods of automatic medical report generation.
\end{abstract}

\section{Introduction}
Deep Convolutional Neural Networks in combination with Recurrent Neural Networks are a common architecture used to automatically generate descriptions of images.
These recent advances have not left other areas such as medical research untouched.
Demner-Fushman et al.~\cite{demner2015preparing} released an anonymized dataset which contains $7470$ chest X-ray images associated with doctors' reports and tag information specifying medical attributes.
However, annotating these domain-specific datasets requires expert-knowledge and cannot be achieved cost-efficiently like more common datasets. In addition, medical data is connected with high privacy concerns and also regulated, e.g., by the Health Insurance Portability and Accountability Act (HIPAA).

Therefore, only a limited amount of data is publicly available. Especially, there is only one public dataset~\cite{demner2015preparing} that connects chest X-ray images with medical reports. In this dataset, there are far more sentences describing normalities than abnormalities. Thus, most machine learning models are biased to generate normal results with higher probability than abnormal results. However, abnormalities are more important and more difficult to detect given the small number of examples. In this work, we address this issue with a new architecture, which can distinguish between generating abnormal or normal sentences.

Furthermore, common machine translation metrics such as BLEU~\cite{papineni2002bleu} may not be the best choice, when even one word - such as `no' - contained in a paragraph can make a huge difference for the indication and findings. Also, calculating these metrics over an imbalanced dataset raises the issue that sentences about normal cases are over-weighted and results in less diversity in the generated reports. We examine these issues of common machine translation metrics when used on a dataset of medical reports such as in our work.

Our contributions: (1) We annotate each sentence of a public dataset with abnormal labels and (2) 
use these labels to train a new hierarchical LSTM with dual word LSTMs combined with an abnormal 
sentence predictor to reduce the data bias. (3) We analyze the correlation between machine 
translation metrics and the variability in generated reports and find that a high score calculated 
over a dataset does not necessarily imply a result to rely on.

\section{Related Work}
\begin{figure}
	\begin{center}
		\begin{subfigure}{.52\linewidth}
				\includegraphics[width=\linewidth]{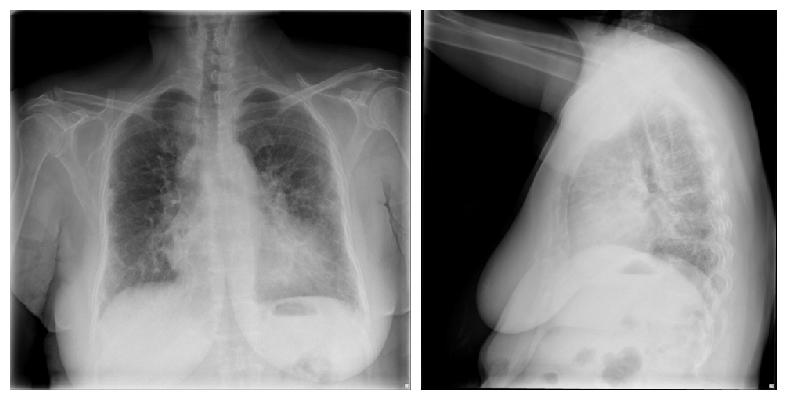}
			\end{subfigure}
			\begin{subfigure}{.395\linewidth}
				\footnotesize{
					\input{paper_imgs/CXR1001.tex}
				}
			\end{subfigure}
	\end{center}	
	\caption{An example from the IU chest X-ray dataset, which shows 
	an abnormal case with findings. We 
	highlight the sentences with our human abnormality annotation, i.e., normal sentences are written in blue and abnormal sentences in green.}
	\label{fig:dataset_example}
\end{figure}
In the field of combining computer vision and machine learning with medical chest X-ray images, Wang et al.~\cite{wang2017chestx}  published the large Chest-Xray14 dataset, which includes a collection of over 100,000 chest X-rays annotated with 14 common thorax diseases. This dataset has been widely~\cite{li2017thoracic, rajpurkar2017chexnet,wang2018tienet} used for predicting and localizing thorax diseases. The disease labels of this dataset were automatically extracted from the doctors' reports. However, the doctors' reports are not available publicly. Demner-Fushman et al.~\cite{demner2015preparing} are the first to release a rather large anonymized dataset consisting of chest X-rays paired with doctors' reports, indications and manually annotated disease labels. We use this dataset in our work.

Automatically generating captions from images is a well-researched topic. Nowadays, most architectures use an encoder-decoder structure, where a Convolutional Neural Network (CNN) is used to encode images into a semantic 
representation and a Recurrent Neural Network (RNN) decoder generates the most likely sentence given this image representation, e.g., ~\cite{vinyals2015show, karpathy2015deep, johnson2016densecap}. 
Krause et al.~\cite{krause2017hierarchical} extended the work by introducing a hierarchical LSTM structure to generate longer sequences for describing an image with a paragraph. 

Jing et al.~\cite{P18-1240} used a hierarchical LSTM to generate a doctor's report with multiple sentences, and use a co-attention mechanism that attends to visual and semantic features, which are generated by medical tags annotated within the Indiana University chest X-ray collection~\cite{demner2015preparing}. Li et al.~\cite{li2018hybrid} describe a hybrid reinforced agent that decides during the process of creating every single sentence if it should be retrieved from a template library or generated in a hierarchical fashion. Instead of a hierarchical model, Xue et al.~\cite{xue2018multimodal} 
use a bidirectional LSTM to encode semantic information of the previously generated sentence as guidance for an attention mechanism to generate an attentive context vector for the current sentence. 
Wang et al.~\cite{wang2018tienet} presented a joint framework, which simultaneously predicts one of 14 diseases and generates a report on the 
Chest-Xray14 dataset. However, the textual annotations are not available to the public as of yet. They use a single LSTM that produces a report conditioned on the previous hidden state, the previously generated words and image features extracted by a CNN.
In a more recent work, Li et al.~\cite{li2019knowledge} use a graph transformer to decompose visual features into an abnormality graph, which is decoded as a template sequence and paraphrased into a generated report.

Our work is based on a hierarchical LSTM structure~\cite{krause2017hierarchical,P18-1240} and introduces an abnormal sentence predictor in combination with a dual word LSTM for separately generating abnormal and normal sentences. In addition, we do not use any templates for sentence generation in contrast to~\cite{li2018hybrid,li2019knowledge}.
\begin{table}
    \begin{center}
    \resizebox{1.0\linewidth}{!}{
    	\begin{tabular}{rrl|rrl} 
    		\toprule
    		rank &  $f$ & sentence & rank &  $f$ & sentence \\
    		\midrule
    		1 & 947 &  no acute cardiopulmonary abnormality & ... & ... & ... \\
    		2 & 698 & the lungs are clear.& 8018 & 1 &  mild right basilar airspace consolidation may ... \\
    		3 & 523 & no pneumothorax. & 8019 & 1 &  calcified granuloma is seen in the left medial... \\
    		4 & 451 & lungs are clear.  & 8020 & 1 & old rib fractures healed.  \\
    		5 & 394 & no acute cardiopulmonary findings.  & 8021 & 1 &  negative for pneumothorax pneumomediastinum or...  \\
    		... & ... & ... & 8022 & 1 &  it is unchanged compared to a for the abdomen ... \\
    		\bottomrule
    	\end{tabular}
    	}
    \end{center}
	\caption{Distinct sentences sorted top-down by their number of appearances $f$.}
	\label{tab:sent_frequency}
\end{table}
\section{Datasets}
\label{sec:datasets}

For our work, we use the Indiana University chest X-ray Collection~\cite{demner2015preparing} (IU chest X-ray dataset), 
which contains 7,470 chest X-Ray images with multiple annotations. These include indication, 
findings, impressions in a textual form and MTI (Medial Text Indexer) encodings. The MTI encodings are 
automatically extracted keywords from the indication and findings. We identify 121 unique MTI 
labels in dataset and use these labels for an additional training signal. Additionally the authors 
manually annotated the images with MEDLINE\textsuperscript{\textregistered} Medical Subject 
Headings\textsuperscript{\textregistered} (MeSH\textsuperscript{\textregistered}). To summarize, this public 
dataset contains 3,955 narrative reports, each associated with MeSH tags and two views of the chest, i.e.,  a Posteroanterior (PA) and a lateral view. We set the doctor's report to be the concatenation of the 
impression and findings similarly to other works~\cite{P18-1240, wang2018tienet}. We show one example from this dataset in Figure~\ref{fig:dataset_example}.

It is very difficult for a machine 
learning model to properly learn the task of generating full paragraphs of doctors' reports from 
this small number of examples. Especially, we notice that most of the reports consist of repeating and very similar sentences, which are of descriptive nature and do not describe abnormalities and diseases. In Table~\ref{tab:sent_frequency}, we list the frequency $f$ of distinct sentences within the doctor's reports, i.e., all sentences that appear at least once in the dataset sorted top-down with most frequent sentences listed on top. We notice a long-tail distribution with abnormal sentences often only occurring with a frequency of $f=1$ within the whole dataset. In fact, 6,290 of the 8,022 distinct sentences have a frequency $f<3$. Machine learning models produce a probability distribution, thus, we always get the most probable doctor's report given the input image. However, most of the images in the dataset depict normal cases and it is difficult to generate accurate reports for abnormal cases. Considering that identifying abnormalities and diseases is the most crucial part in this problem domain, we want to address the data bias problems for chest X-ray image report generation.
\section{Dual Word LSTM Medical Image Report Generation}

\begin{figure}
	\centering
	\includegraphics[width=0.8\columnwidth]{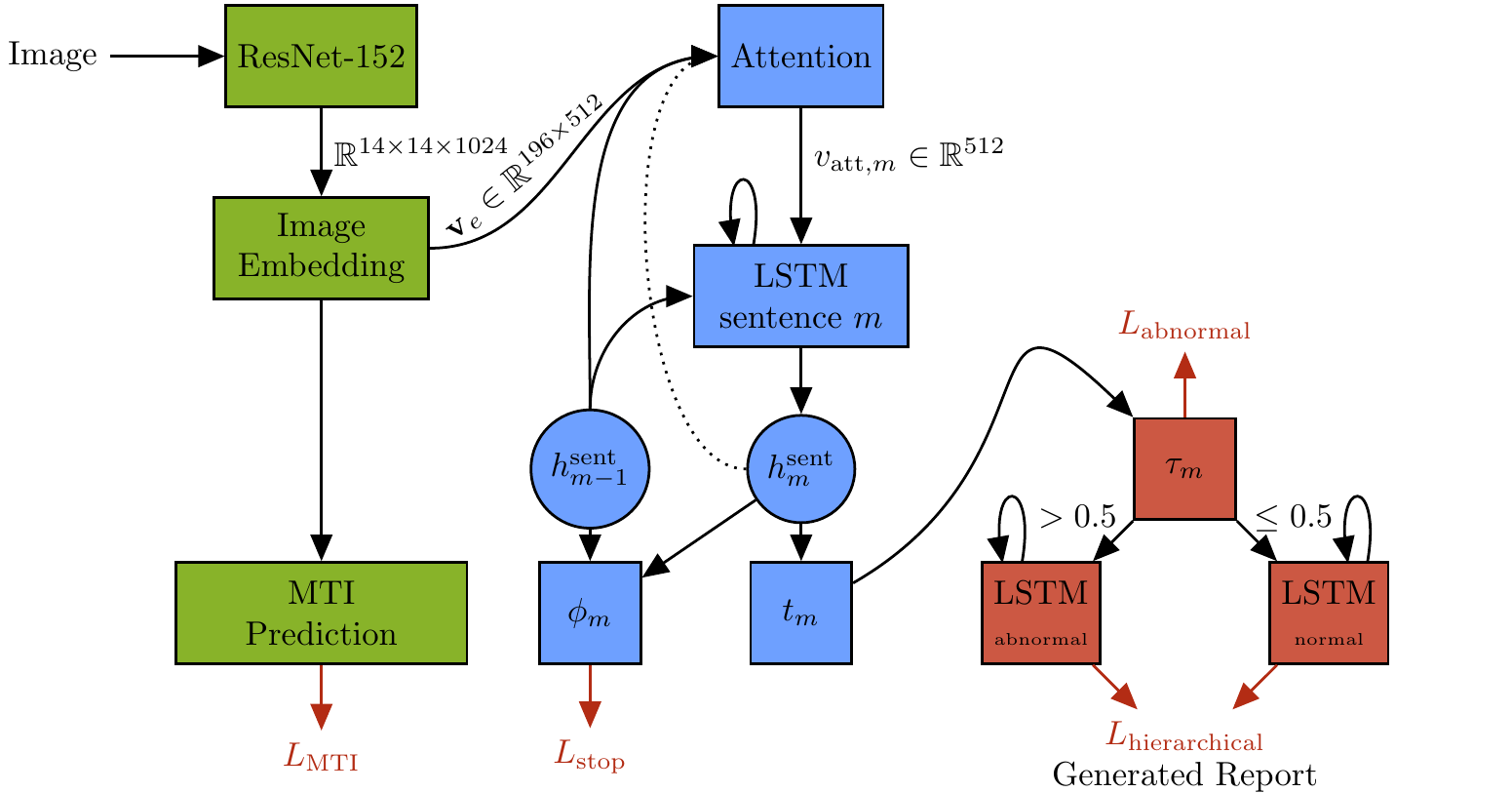}
	\caption{Our dual word LSTM model. The green boxes show the CNN image encoder, the image embedding 
	and the MTI tag prediction. The blue part depicts the sentence LSTM, i.e., the topic 
	generator $t_m$ and the stop prediction $\phi_m$. The red part shows an abnormal 
	sentence predictor ($\tau_m$) and two word LSTMs for generating abnormal and normal sentences, respectively.} 
	\label{fig:model}	
\end{figure}
We depict our model architecture in Figure~\ref{fig:model}. The input to our model are single images, i.e., either the lateral view or PA view of a chest X-ray image. We use the \textit{res4b35} feature map of the ResNet-152~\cite{he2016deep} $\mathbf{\tilde{v}} \in 
\mathbb{R}^{14\times14\times1024}$ as our image features. $\mathbf{v}_e \in \mathbb{R}^{14\times14\times512}$  embeds these images features into a lower dimensional space for further use.
It is reshaped into a feature map of shape $\mathbb{R}^{196\times 512}$ enabling a soft attention mechanism to attend to 196 different spatial locations. Unless otherwise noted, all embedding and hidden dimensions are set to $512$ in our model.  

\subsection{Hierachical Generation with Dual Word LSTMs}

Even though LSTMs were designed to combat the issue of forgetting long-term dependencies, they still have problems keeping information for very long time-periods, e.g., over multiple sentences. Krause et al. \cite{krause2017hierarchical} address this problem by splitting the generation into a hierarchical LSTM, which consists of two independent LSTMs. The sentence LSTM's sole purpose is to generate topic vectors, which in turn are used for the initialization of the word LSTM. The word LSTM then generates a single sentence conditioned on the topic vector. Jing et al.~\cite{P18-1240} extend the hierachical LSTM for generating medical reports from chest X-ray images. We also use a hierarchical concept with an architecture that differs from Jing et al.~\cite{P18-1240}. For example, we add a multi-task learning objective on MTI tags (see Section~\ref{sec:learn_obj}) and do not use the Co-Attention mechanism in our model.
\paragraph{Sentence LSTM}
We initialize the sentence LSTM on image features extracted by the encoder CNN. However, we use a soft attention mechanism $\mathbf{v}_m = f_{\textrm{att}}(\mathbf{v}_e, h^{\textrm{sent}}_{m-1})$ to attend to different spatial areas within the feature map conditioned on the sentence LSTM's hidden state $h^{\textrm{sent}}_{m-1}$ of the preceding sentence. In subsequent sentences, we use the corresponding preceding hidden state, which we depicted by the dotted arrow in Figure~\ref{fig:model}. In order to generate the topic vector for sentence $m$, we apply the sentence LSTM to the attentive image features $\mathbf{v}_m$ to get an intermediate hidden state $ h^{\textrm{sent}}_{m}$ for the current sentence and feed it through a 
fully-connected layer
\begin{equation}
\mathbf{t}_m = \text{relu}(\mathbf{W_{\textrm{sent}}}h^{\textrm{sent}}_m)
\end{equation}
to generate a topic vector, where $\mathbf{W_{\textrm{sent}}} \in R^{\textrm{word embedding dim}}$.

\paragraph{Stop Prediction} 
We also use the sentence LSTM's current and previous hidden state to predict if we should continue generating sentences ($z_m=0$) or 
stop generating them ($z_m=1$). The stop prediction ($\phi_m$) is a fully-connected layer 
\begin{equation}
\phi_m=\mathbf{W}_{\textrm{stop}}\tanh(\mathbf{W}_{\textrm{stop},m-1}h^{\textrm{sent}}_{m-1}+ 
\mathbf{W}_{\textrm{stop},m}h^{\textrm{sent}}_{m}),
\end{equation}
where $\mathbf{W}_{\textrm{stop}}$, $\mathbf{W}_{\textrm{stop},m-1}$ and $\mathbf{W}_{\textrm{stop},m}$ are parameter matrices. We train the stop prediction with a sigmoid cross-entropy loss $L_{\text{stop}} = - \sum_{m=0}^{M-1} z_m \cdot \log(\sigma(\phi_m)) + (1-z_m) \cdot \log(1-\sigma(\phi_m))$,
where $\sigma$ is the Sigmoid function and $M$ is the number of sentences in the current paragraph.

\paragraph{Dual Word LSTMs} 
A word LSTM is trained to maximize the probability of 
predicting the ground-truth word $w_{m,t}$ at timestep $t$ of sentence $m$. The hierachical LSTM softmax cross-entropy loss is then defined by
\begin{equation}
\label{eq:hierarchical_loss}
L_{\textrm{hierachical}} = \sum_{m=0}^{M-1} \sum_{t=0}^{N_m -1} w_{m,t} \log 
(o^{\textrm{word}}_{m,t}),
\end{equation}
where $N_m$ are the number of words in sentence $m$ and $o^{\textrm{word}}_{m,t}$ the output of the 
word LSTM at timestep $t$ of sentence $m$. The input to time step $t$ is the embedded ground truth 
word $\mathbf{W}_e w_{m, t-1}$, where  $\mathbf{W}_e$ is the word embedding matrix.

Depending on whether the current sentence is of type abnormal or normal, we train a different set of word LSTM parameters. In other words, we have an abnormal word LSTM and a normal word LSTM, which are trained when the label of the current sentence is abnormal and normal, respectively. In practice, we set the loss weights for the current sentence to 1 in the abnormal word LSTM and to 0 in normal word LSTM. In the case of a normal sentence, we set the loss weights inversely. During inference phase, we use the prediction of the abnormal sentence prediction module (see Section~\ref{sec:abnpred}) to decide whether we want to use the generated sentence from the abnormal word LSTM or the normal word LSTM. We then concatenate sentences from both the abnormal word LSTM and the normal word LSTM into our final paragraph.

\subsection{Abnormal Sentence Prediction}
\label{sec:abnpred}
As we already argued in Section~\ref{sec:datasets}, the dataset consists of many distinct normal sentences, but only few different sentences exist that describe abnormalities. We integrate an abnormality prediction module, which tries to infer whether the semantic meaning of topic vector $\mathbf{t}_m$ does describe an abnormality or not. We use a fully-connected layer $\tau_m$ with one output neuron to predict the probability for a sentence to be abnormal or not. We train the fully-connected layer with a sigmoid cross-entropy loss $L_{\text{abnormal}}$.

We manually annotated the IU chest X-ray dataset for every sentence within the ground-truth 
paragraph of every sample in the training dataset. Two annotators labeled whether a sentence is an 
abnormal case or not with the help of the provided MeSH tags. In addition, we also implemented a 
method for automatically annotating the sentences by comparing word embedding distances against 
MeSH tag embeddings although we use manual annotations for training.  We use 
Word2Vec~\cite{mikolov2013efficient, mikolov2013distributed} embeddings trained on Pubmed and 
Wikipedia~\cite{moen2013distributional} which can reduce human efforts when the dataset is scaled 
up.


\subsection{Learning Objective}
\label{sec:learn_obj}

We use the global average pool of the image embedding $\mathbf{\hat{v}}\in\mathbb{R}^{512}=\textrm{avg\_pool}(\mathbf{v}_e)$ for predicting the MTI annotations. As it is common in multi-label classification, we use the sigmoid cross-entropy loss function $L_{\textrm{MTI}}$ appended to a fully-connected layer with one output neuron for every distinct MTI label.
For our experiments, we optimize the total loss
\begin{equation}
L_{\textrm{total}} = \lambda_{\textrm{stop}} \cdot L_{\textrm{stop}} + 
\lambda_{\textrm{hierarchical}} \cdot L_{\textrm{hierarchical}} + \lambda_{\textrm{abnormal}} 
\cdot L_{\textrm{abnormal}} + \lambda_{\textrm{MTI}} \cdot L_{\textrm{MTI}},
\end{equation}
where $\lambda_{(\cdot)}$ are the weighting factors for each loss. $\lambda_{\textrm{MTI}}$ is set 
to $10$ and $\lambda_{\textrm{hierarchical}}$, $\lambda_{\textrm{stop}}$ and 
$\lambda_{\textrm{abnormal}}$ are set to $1$. 
We set $\lambda_{\textrm{abnormal}}$ to $0$ for experiments in which we disable the dual LSTM 
approach, i.e., we only use a single word LSTM similar to Jing et al.~\cite{P18-1240}.
In this case, we also calculate 
$L_{\textrm{hierarchical}}$ with only one word LSTM. When using the abnormal and normal word LSTMs, 
$L_{\textrm{hierarchical}}$ is the sum of the two individual word LSTM losses, i.e., 
$o^{\textrm{word}}_{m,t}$ from Equation~\ref{eq:hierarchical_loss} is the output of the abnormal or 
normal word LSTM, depending on whether the ground-truth annotation is abnormal or normal.

We train the image embedding layer with both the hierachical LSTM and the MTI predictor, so the 
captioning task can benefit from our multi-task 
loss function. We use the Adam~\cite{kingma2014adam} optimizer with a base learning rate of 
$\eta=5\cdot10^{-4}$ and do not use learning rate decay. We train 
for up to $250$ epochs and use a batch size of $16$.

\begin{table}
	\begin{center}
\resizebox{\linewidth}{!}{
\begin{tabular}{@{}lccccccc@{}}
\toprule
Model & B-1 & B-2 & B-3 & B-4 & Cider & Meteor & Rouge-L \\ \midrule
CNN-RNN~\cite{vinyals2015show} & 31.9 (33.3) & 19.8 (20.5) & 13.3 (13.6) & 9.4 (9.4) & 29.1 (30.6) & 13.5 (14.5) & 26.8 (27.2) \\
CoAtt\textsuperscript{*}~\cite{P18-1240} & --- (45.5) & --- (28.8) & --- (20.5) & --- (15.4) & --- (27.7) & --- (---) & --- (36.9) \\
KERP\textsuperscript{*}~\cite{li2019knowledge} & --- (48.2) & --- (32.5) & --- (22.6) & --- (16.2) & --- (28.0) & --- (---) & --- (33.9) \\
\midrule
HLSTM & 36.4 (\textbf{37.6}) & 23.2 (23.8) & 16.1 (16.3) & 11.4 (11.4) & 29.1 (29.3) & 15.5 (15.7) & 30.6 (30.2) \\
HLSTM+att & 35.1 (36.6) & 22.8 (23.4) & 16.1 (16.4) & 11.6 (11.7) & 34.3 (32.3) & 14.9 (15.6) & 29.7 (29.9) \\
HLSTM+Dual & 35.2 (35.8) & 22.8 (23.1) & 15.9 (16.0) & 11.3 (11.2) & 34.8 (32.2) & 14.6 (15.1) & 29.5 (29.6) \\
HLSTM+att+Dual & 35.7 (37.3) & 23.3 (\textbf{24.6}) & 16.5 (\textbf{17.5}) & 11.8 (\textbf{12.6}) & 34.0 (\textbf{35.9}) & 15.6 (\textbf{16.3}) & 31.3 (\textbf{31.5}) \\ \bottomrule
\end{tabular}
}
	\end{center}
	\caption{Results (in \%) on the validation and (test) set calculated with common machine translation metrics. B-n stands for BLEU-n which uses up to $n$-grams. We selected the model configuration and hyperparameters based on the validation set. \textit{HLSTM} / \textit{HLSTM+att} are our hierarchical LSTM implementations similar to \cite{krause2017hierarchical,P18-1240}, and are evaluated on our dataset splits. \textit{\dots+Dual} are our models.
	\textsuperscript{*}Scores taken from~\cite{li2019knowledge}, who used a different dataset split.}	
	\label{tab:valscores}
\end{table}
\section{Experiments and Evaluation}
In the following, we present an evaluation study and results generated by our hierarchical models with dual word LSTMs, \textit{HLSTM+Dual} and 
\textit{HLSTM+att+Dual}. We choose to compare against the CNN-RNN~\cite{vinyals2015show} baseline 
which we trained ourselves on our train split. We also compared our model against the scores 
reported in CoAtt~\cite{P18-1240} and KERP~\cite{li2019knowledge}. These models were pretrained on 
a non-public dataset of chest X-ray images 
with Chinese reports, which were collected by a professional medical institution for health checking~\cite{li2018hybrid}. KERP~\cite{li2019knowledge} uses templates which is different in comparison to end-to-end generation approaches.
However, since these methods were evaluated on a different dataset split, the scores are not directly comparable to ours. We therefore implemented hierarchical LSTM baselines similar to~\cite{krause2017hierarchical} which were referred to in CoAtt~\cite{P18-1240}. These hierarchical LSTM baselines with and without an attention mechanism are named \textit{HLSTM} and \textit{HLSTM+att} in our paper, and are evaluated on our dataset split.
As we do not have access to the CX-CHR dataset from~\cite{li2018hybrid}, we did not employ any pretraining of the feature extractor network.

\subsection{Model Selection}
\begin{figure}
\begin{center}
\begin{tabular}{C{.4\linewidth}  L{.59\linewidth}}
        \includegraphics[width=\linewidth]{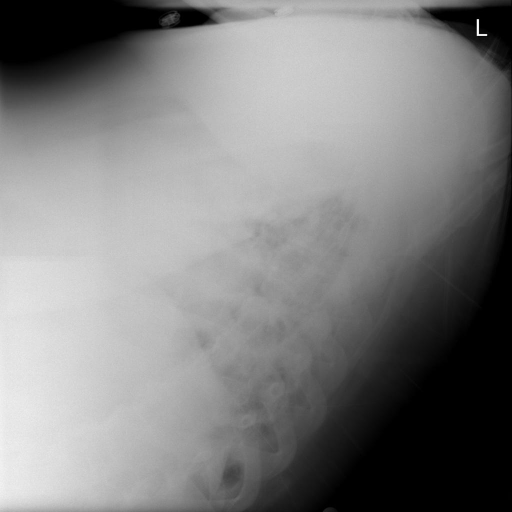} & \footnotesize{\input{paper_imgs/CXR2194.F2.tex}}
\end{tabular}
\end{center}
	\caption{Examples of generated paragraphs with our model \textit{HLSTM+Dual} vs. \textit{HLSTM} in comparison with the ground-truth paragraph.}
	\label{fig:generated_example}
\end{figure}

We choose our dataset split by randomly 
shuffling the dataset and splitting it into a train, validation and test set with a ratio of $0.9$, 
$0.05$ and $0.05$, respectively. We make sure that images of an individual patient are only present 
in either one of train, validation or test set.
We use the validation set for selection of hyperparameters and architectural decisions. In 
practice, we select the best model checkpoint based on two criteria. First, we calculate metrics 
such as BLEU-n twice per training epoch. Second, we also calculate the number of distinct 
sentences generated over the whole validation dataset for each sentence index $m$ within a 
paragraph. 
We choose our final models by calculating these criteria over the validation set. (1) The first sentences $m=0$ over the whole validation should at least comprise of $4$ distinct sentences. (2) We select the model with the highest BLEU-4 score.  We depict the scores on the validation set in 
Table~\ref{tab:valscores}. We report the scores on the held-out test set in brackets and show paragraphs generated by \textit{HLSTM} and \textit{HLSTM+Dual} in Figure~\ref{fig:generated_example}.
\begin{figure}
	\centering
	\includegraphics[width=\columnwidth]{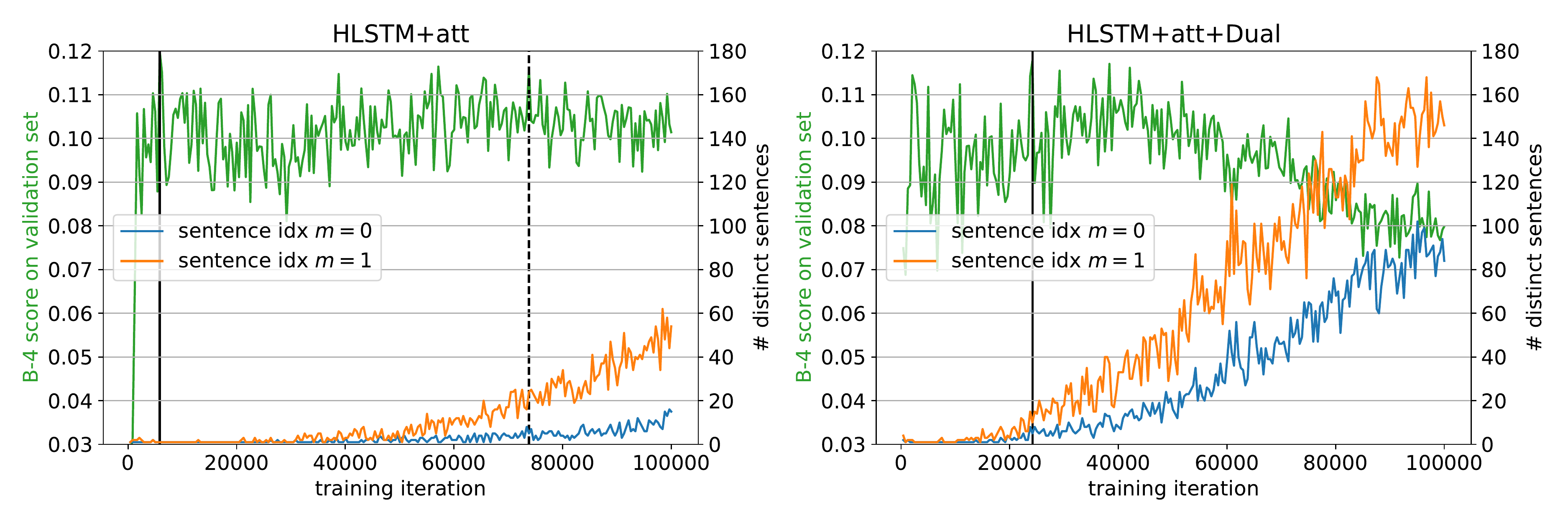}
	\caption{The number of distinct sentences of sentence $m=0$ and $m=1$ plotted against the BLEU-4 
		validation score over the course of training for \textit{HLSTM+att} and 
		\textit{HLSTM+att+Dual}. The solid line represents the training iteration with the maximum 
		BLEU-4 score and the dashed line our selected model.}
	\label{fig:sent_stats}
\end{figure}
\subsection{Analysis on Evaluation Scores and Distinct Sentences}
\label{sec:res_distinct_sents}
We observed a severe disadvantage in solely using scores such as BLEU-n as the evaluation criteria. 
As we mentioned before, we calculated the number of distinct sentences per sentence index $m$ 
within a generated paragraph for each validation point. We noticed that high scores do not 
necessarily imply a high variability in generated sentences. Most notably, the highest scores can 
sometimes be observed when there are only $1$ or $2$ distinct sentences per sentence index resulting in 
very few 
different paragraphs. In Figure~\ref{fig:sent_stats}, we show the number of distinct sentences for 
sentence indices $m=0$ and $m=1$ compared to the BLEU-4 score over the course of training. We see 
that the 
score of model \textit{HLSTM+att} stays in the same limited range over the course of training. For example, it has a 
higher score even though it generates the very same paragraph for every sample in the validation 
set at training iteration $5838$ (visualized by the vertical black line) than at training iteration 
$73809$ 
(visualized by dashed line). For the model \textit{HLSTM+att+Dual}, we see that the score drops 
as more distinct sentences are generated. For this model, we also see that there is much more 
variability of sentences from the beginning on and also far more distinct sentences are generated 
in contrast to only using a single word LSTM.
\begin{table}
	\begin{center}
		\resizebox{0.8\linewidth}{!}{
        \begin{tabular}{lcccccccccc}
        \toprule
        \diagbox{Model}{$m$} & $0$ & $1$ & $2$ & $3$ & $4$ & $5$ & $6$ & $7$ & $8$ & $9$ \\ \midrule
        GT & 1216 & 1540 & 1586 & 1549 & 1378 & 1086 & 725 & 477 & 278 & 171 \\
        CNN-RNN~\cite{vinyals2015show} & 12 & 19 & 17 & 23 & 19 & 8 & 0 & 0 & 0 & 0 \\ \midrule
        HLSTM+att & 5 & 24 & 24 & 33 & 25 & 31 & 23 & 14 & 9 & 3 \\
        HLSTM & 4 & 13 & 12 & 18 & 25 & 22 & 15 & 14 & 11 & 4 \\
        HLSTM+Dual & 8 & 28 & 36 & 45 & 32 & 17 & 2 & 0 & 0 & 0 \\
        HLSTM+att+Dual & 5 & 10 & 7 & 8 & 8 & 4 & 1 & 0 & 0 & 0 \\ \bottomrule
        \end{tabular}
		}
	\end{center}
	\caption{Number of distinct sentences in the ground-truth (GT) and generated on the validation 
		split of the dataset per 
		sentence index $m \in [0,9]$ within the generated paragraph.}
	\label{tab:num_sents}
\end{table}
If we look at the number of distinct sentences generated per sentence index $m$ in our chosen 
models compared to the ground-truth in Table~\ref{tab:num_sents}, we still see a huge gap. 
Note that paragraphs with mostly one distinct sentence per sentence index do not have 
additional benefit, since they are not dependent on the input image.
Considering that many sentences within the ground-truth only differ slightly but have a synonymous 
meaning, we find that results which do not possibly have the maximum score but a higher variability 
in generated paragraphs describe the input images in a better way. Thus, we also use a minimum 
threshold of distinct sentences as one stopping criterion.

\subsection{Dual Word LSTM with Abnormal Sentence Predictor}
The test scores of our models and the baselines are presented in Table~\ref{tab:valscores} (in brackets). Over all evaluation metrics, our \textit{HLSTM+att+Dual} model has the most improvement on Cider~\cite{vedantam2015cider}, which is designed for evaluating image descriptions, uses human consensus and considers the TF-IDF for weighting n-grams. This implies that our \textit{HLSTM+att+Dual} model can catch more distinct n-grams in the reference paragraph. In addition, our \textit{HLSTM+att+Dual} model is consistently better than other baselines in multi-gram BLEU, Meteor and Rouge-L, indicating that the relevance is not sacrificed while distinctiveness is increased.
In addition, we also compared our models with the dual word LSTM from Section~\ref{sec:abnpred} 
against the vanilla HLSTM model inspired by Jing et al.~\cite{P18-1240}. As we already mentioned in 
Section~\ref{sec:res_distinct_sents}, the number of distinct sentences per each sentence index 
starts to grow more rapidly when using two word LSTMs, which can be seen in the right part of 
Figure~\ref{fig:sent_stats} when comparing it to the \textit{HLSTM+att} model on the left. We can 
also see that generating more distinct sentences does not account for better scores. However, when 
looking at the validation and test scores in Table~\ref{tab:valscores} the dual word LSTM models 
often have higher scores than the single word LSTM models.
\begin{table}
	\begin{center}
		\resizebox{\linewidth}{!}{
		\begin{tabular}{@{}lccccccc@{}}
        \toprule
        Model & B-1 & B-2 & B-3 & B-4 & Cider & Meteor & Rouge-L \\ \midrule
        HLSTM+att & 30.9 (44.4) & 19.0 (30.1) & 12.9 (21.8) & 9.1 (15.8) & 25.9 (42.6) & 12.8 (22.2) & 25.0 (38.6) \\
        HLSTM & 32.3 (43.5) & 19.4 (29.7) & 12.8 (21.3) & 8.8 (15.3) & 24.6 (37.1) & 13.2 (21.7) & 25.8 (38.2) \\
        HLSTM+Dual & \textbf{32.8} (41.2) & \textbf{20.6} (28.1) & \textbf{14.0} (20.0) & \textbf{9.8} (13.9) & \textbf{30.1} (31.8) & 13.2 (19.5) & 26.1 (36.0) \\
        HLSTM+att+Dual & 31.8 (\textbf{46.9}) & 19.8 (\textbf{33.5}) & 13.5 (\textbf{24.9}) & 9.7 (\textbf{18.3}) & 28.4 (\textbf{49.5}) & \textbf{13.5} (\textbf{22.8}) & \textbf{26.9} (\textbf{39.9}) \\ \bottomrule
        \end{tabular}
		}
\end{center}	
	\caption{Final results (in \%) on the held-out test-set of the dataset for ABNORMAL and (NORMAL) images 
	only.}
\label{tab:abnormal_normal_scores}
\end{table}

In Table~\ref{tab:abnormal_normal_scores}, we report scores on the held-out test set for abnormal 
as well as normal (in brackets) images. 
The best-performing model for both normal and abnormal images was one of our dual models.
The results also indicate that the performance is best on normal images and so effort should be given to further improve performance on abnormal images.
\section{Conclusion and Future Work}
In our work, we presented a hierarchical LSTM architecture expanded by a dual word LSTM. Paired 
with an abnormality prediction module, we introduced dual word LSTMs, which are responsible for generating abnormal 
and normal sentences, respectively. 
 
We then examined the correlation between the BLEU-n metrics and the number of distinct sentences generated 
by our model and observed that common evaluation metrics such as BLEU-4 do not necessarily imply a good 
evaluation criteria for multi-sentence medical reports, i.e., for one of our models the highest 
score was produced by only generating the same paragraph for every input image. In addition, note that the dual word LSTM can help to increase the number of distinct sentences faster when selecting a corresponding stopping criterion.

In the future, we want to focus on working on a metric more suitable for the critical area of medical report generation from images and addressing abnormal indications and findings, since their performance is worse than those of normal indications and findings.

\section{Acknowledgments}

This work was done during Philipp Harzig's internship at FX Palo Alto Laboratory. He thanks 
his colleagues from FXPAL for the collaboration, advice and for providing an open and inspiring 
research environment. We also thank Eric Rosenberg for helping annotate the ground-truth sentences 
with abnormal/normal labels.

\bibliography{egbib}

\begin{thebibliography}{20}
\providecommand{\natexlab}[1]{#1}
\providecommand{\url}[1]{\texttt{#1}}
\expandafter\ifx\csname urlstyle\endcsname\relax
  \providecommand{\doi}[1]{doi: #1}\else
  \providecommand{\doi}{doi: \begingroup \urlstyle{rm}\Url}\fi

\bibitem[Demner-Fushman et~al.(2015)Demner-Fushman, Kohli, Rosenman, Shooshan,
  Rodriguez, Antani, Thoma, and McDonald]{demner2015preparing}
Dina Demner-Fushman, Marc~D Kohli, Marc~B Rosenman, Sonya~E Shooshan, Laritza
  Rodriguez, Sameer Antani, George~R Thoma, and Clement~J McDonald.
\newblock Preparing a collection of radiology examinations for distribution and
  retrieval.
\newblock \emph{Journal of the American Medical Informatics Association},
  23\penalty0 (2):\penalty0 304--310, 2015.

\bibitem[He et~al.(2016)He, Zhang, Ren, and Sun]{he2016deep}
Kaiming He, Xiangyu Zhang, Shaoqing Ren, and Jian Sun.
\newblock Deep residual learning for image recognition.
\newblock In \emph{Proceedings of the IEEE conference on computer vision and
  pattern recognition}, pages 770--778, 2016.

\bibitem[Jing et~al.(2018)Jing, Xie, and Xing]{P18-1240}
Baoyu Jing, Pengtao Xie, and Eric Xing.
\newblock On the automatic generation of medical imaging reports.
\newblock In \emph{Proceedings of the 56th Annual Meeting of the Association
  for Computational Linguistics (Volume 1: Long Papers)}, pages 2577--2586.
  Association for Computational Linguistics, 2018.
\newblock URL \url{http://aclweb.org/anthology/P18-1240}.

\bibitem[Johnson et~al.(2016)Johnson, Karpathy, and
  Fei-Fei]{johnson2016densecap}
Justin Johnson, Andrej Karpathy, and Li~Fei-Fei.
\newblock Densecap: Fully convolutional localization networks for dense
  captioning.
\newblock In \emph{Proceedings of the IEEE Conference on Computer Vision and
  Pattern Recognition}, pages 4565--4574, 2016.

\bibitem[Karpathy and Fei-Fei(2015)]{karpathy2015deep}
Andrej Karpathy and Li~Fei-Fei.
\newblock Deep visual-semantic alignments for generating image descriptions.
\newblock In \emph{Proceedings of the IEEE conference on computer vision and
  pattern recognition}, pages 3128--3137, 2015.

\bibitem[Kingma and Ba(2014)]{kingma2014adam}
Diederik~P Kingma and Jimmy Ba.
\newblock Adam: A method for stochastic optimization.
\newblock \emph{arXiv preprint arXiv:1412.6980}, 2014.

\bibitem[Krause et~al.(2017)Krause, Johnson, Krishna, and
  Fei-Fei]{krause2017hierarchical}
Jonathan Krause, Justin Johnson, Ranjay Krishna, and Li~Fei-Fei.
\newblock A hierarchical approach for generating descriptive image paragraphs.
\newblock In \emph{Computer Vision and Pattern Recognition (CVPR), 2017 IEEE
  Conference on}, pages 3337--3345. IEEE, 2017.

\bibitem[Li et~al.(2018)Li, Liang, Hu, and Xing]{li2018hybrid}
Christy~Y Li, Xiaodan Liang, Zhiting Hu, and Eric~P Xing.
\newblock Hybrid retrieval-generation reinforced agent for medical image report
  generation.
\newblock \emph{arXiv preprint arXiv:1805.08298}, 2018.

\bibitem[Li et~al.(2019)Li, Liang, Hu, and Xing]{li2019knowledge}
Christy~Y Li, Xiaodan Liang, Zhiting Hu, and Eric~P Xing.
\newblock Knowledge-driven encode, retrieve, paraphrase for medical image
  report generation.
\newblock \emph{AAAI Conference on Artificial Intelligence}, 2019.

\bibitem[Li et~al.(2017)Li, Wang, Han, Xue, Wei, Li, and Li]{li2017thoracic}
Zhe Li, Chong Wang, Mei Han, Yuan Xue, Wei Wei, Li-Jia Li, and F~Li.
\newblock Thoracic disease identification and localization with limited
  supervision.
\newblock \emph{arXiv preprint arXiv:1711.06373}, 2017.

\bibitem[Mikolov et~al.(2013{\natexlab{a}})Mikolov, Chen, Corrado, and
  Dean]{mikolov2013efficient}
Tomas Mikolov, Kai Chen, Greg Corrado, and Jeffrey Dean.
\newblock Efficient estimation of word representations in vector space.
\newblock \emph{Proceedings of Workshop at ICLR, 2013}, 2013{\natexlab{a}}.

\bibitem[Mikolov et~al.(2013{\natexlab{b}})Mikolov, Sutskever, Chen, Corrado,
  and Dean]{mikolov2013distributed}
Tomas Mikolov, Ilya Sutskever, Kai Chen, Greg~S Corrado, and Jeff Dean.
\newblock Distributed representations of words and phrases and their
  compositionality.
\newblock In \emph{Advances in neural information processing systems}, pages
  3111--3119, 2013{\natexlab{b}}.

\bibitem[Moen and Ananiadou(2013)]{moen2013distributional}
SPFGH Moen and Tapio Salakoski2~Sophia Ananiadou.
\newblock Distributional semantics resources for biomedical text processing.
\newblock In \emph{Proceedings of the 5th International Symposium on Languages
  in Biology and Medicine, Tokyo, Japan}, pages 39--43, 2013.

\bibitem[Papineni et~al.(2002)Papineni, Roukos, Ward, and
  Zhu]{papineni2002bleu}
Kishore Papineni, Salim Roukos, Todd Ward, and Wei-Jing Zhu.
\newblock Bleu: a method for automatic evaluation of machine translation.
\newblock In \emph{Proceedings of the 40th annual meeting on association for
  computational linguistics}, pages 311--318. Association for Computational
  Linguistics, 2002.

\bibitem[Rajpurkar et~al.(2017)Rajpurkar, Irvin, Zhu, Yang, Mehta, Duan, Ding,
  Bagul, Langlotz, Shpanskaya, et~al.]{rajpurkar2017chexnet}
Pranav Rajpurkar, Jeremy Irvin, Kaylie Zhu, Brandon Yang, Hershel Mehta, Tony
  Duan, Daisy Ding, Aarti Bagul, Curtis Langlotz, Katie Shpanskaya, et~al.
\newblock Chexnet: Radiologist-level pneumonia detection on chest x-rays with
  deep learning.
\newblock \emph{arXiv preprint arXiv:1711.05225}, 2017.

\bibitem[Vedantam et~al.(2015)Vedantam, Lawrence~Zitnick, and
  Parikh]{vedantam2015cider}
Ramakrishna Vedantam, C~Lawrence~Zitnick, and Devi Parikh.
\newblock Cider: Consensus-based image description evaluation.
\newblock In \emph{Proceedings of the IEEE conference on computer vision and
  pattern recognition}, pages 4566--4575, 2015.

\bibitem[Vinyals et~al.(2015)Vinyals, Toshev, Bengio, and
  Erhan]{vinyals2015show}
Oriol Vinyals, Alexander Toshev, Samy Bengio, and Dumitru Erhan.
\newblock Show and tell: A neural image caption generator.
\newblock In \emph{Proceedings of the IEEE conference on computer vision and
  pattern recognition}, pages 3156--3164, 2015.

\bibitem[Wang et~al.(2017)Wang, Peng, Lu, Lu, Bagheri, and
  Summers]{wang2017chestx}
Xiaosong Wang, Yifan Peng, Le~Lu, Zhiyong Lu, Mohammadhadi Bagheri, and
  Ronald~M Summers.
\newblock Chestx-ray8: Hospital-scale chest x-ray database and benchmarks on
  weakly-supervised classification and localization of common thorax diseases.
\newblock In \emph{Computer Vision and Pattern Recognition (CVPR), 2017 IEEE
  Conference on}, pages 3462--3471. IEEE, 2017.

\bibitem[Wang et~al.(2018)Wang, Peng, Lu, Lu, and Summers]{wang2018tienet}
Xiaosong Wang, Yifan Peng, Le~Lu, Zhiyong Lu, and Ronald~M Summers.
\newblock Tienet: Text-image embedding network for common thorax disease
  classification and reporting in chest x-rays.
\newblock In \emph{Proceedings of the IEEE Conference on Computer Vision and
  Pattern Recognition}, pages 9049--9058, 2018.

\bibitem[Xue et~al.(2018)Xue, Xu, Long, Xue, Antani, Thoma, and
  Huang]{xue2018multimodal}
Yuan Xue, Tao Xu, L~Rodney Long, Zhiyun Xue, Sameer Antani, George~R Thoma, and
  Xiaolei Huang.
\newblock Multimodal recurrent model with attention for automated radiology
  report generation.
\newblock In \emph{International Conference on Medical Image Computing and
  Computer-Assisted Intervention}, pages 457--466. Springer, 2018.

\end{thebibliography}
\end{document}